\documentclass[]{amap}

\usepackage[toc,page,header]{appendix}
\usepackage{minitoc}
\usepackage{solarized-light}
\usepackage{booktabs}
\usepackage{multirow}
\usepackage{graphicx}
\usepackage{array}
\usepackage{siunitx,array}
\usepackage[table]{xcolor}
\usepackage{makecell,array,booktabs}
\usepackage{makecell}
\usepackage{framed}

\usepackage{bbding}
\usepackage{hyperref}
\usepackage{amsmath}
\usepackage{amsfonts}
\usepackage{placeins}
\usepackage[colorinlistoftodos]{todonotes}
\usepackage{longtable}
\usepackage{hhline}
\usepackage{fancyvrb}
\usepackage{graphicx}
\usepackage[table]{xcolor}
\usepackage{booktabs}
\usepackage{float}
\usepackage{fvextra}
\usepackage{CJKutf8}
\usepackage{multicol}
\usepackage{float}
\usepackage{placeins}
\usepackage{cleveref}
\usepackage{tablefootnote}
\usepackage{threeparttable}
\usepackage{tabularx}
\usepackage{mdframed}
\usepackage{subcaption}
\usepackage[usestackEOL]{stackengine}
\usepackage[numbers]{natbib}
\newcommand{\commentout}[1]{}
\renewcommand{\paragraph}[1]{\noindent\textbf{#1.}\hspace*{1em}}
\usepackage{enumitem}
\usepackage{hyperref}
\setlist[itemize]{leftmargin=15pt}
\usepackage{siunitx,array}
\usepackage{tabularx}
\usepackage{adjustbox}
\usepackage{arydshln}
\usepackage{pifont}
\usepackage{bbding}
\definecolor{ampblue}{rgb}{0.82, 0.88, 0.94}
\usepackage[table]{xcolor} 
\usepackage{array} 
\usepackage[dvipsnames]{xcolor}

\RequirePackage{xspace}
\makeatletter
\DeclareRobustCommand\onedot{\futurelet\@let@token\@onedot}
\def\@onedot{\ifx\@let@token.\else.\null\fi\xspace}

\makeatother

\usepackage{xcolor}

\definecolor{abot1}{HTML}{0185FE}
\definecolor{abot2}{HTML}{0185FE}
\definecolor{abot3}{HTML}{0185FE}
\definecolor{abot4}{HTML}{0185FE}
\definecolor{abot5}{HTML}{FB8C00}
\definecolor{abot6}{HTML}{FB8C00}
\definecolor{abot7}{HTML}{FB8C00}

\title{ABot-Claw: A Foundation for Persistent, Cooperative, and Self-Evolving Robotic Agents}

\author{AMAP CV Lab}
\vspace{-11pt}

\contribution{See \hyperref[sec:contributions]{Contributions} section for a full author list.}

\abstract{

Current embodied intelligent systems still face a substantial gap between high-level reasoning and low-level physical execution in open-world environments. Although Vision-Language-Action (VLA) models provide strong perception and intuitive responses, their open-loop nature limits long-horizon performance. Agents incorporating System 2 cognitive mechanisms improve planning, but usually operate in closed sandboxes with predefined toolkits and limited real-system control. OpenClaw provides a localized runtime with full system privileges, but lacks the embodied control architecture required for long-duration, multi-robot execution. We therefore propose \textbf{ABot-Claw}, an embodied extension of OpenClaw that integrates: 1) a \textbf{unified embodiment interface with capability-driven scheduling} for heterogeneous robot coordination; 2) a \textbf{visual-centric cross-embodiment multimodal memory} for persistent context retention and grounded retrieval; and 3) a \textbf{critic-based closed-loop feedback mechanism} with a generalist reward model for online progress evaluation, local correction, and replanning. With a decoupled architecture spanning the OpenClaw layer, shared service layer, and robot embodiment layer, ABot-Claw enables real-world interaction, closes the loop from natural language intent to physical action, and supports progressively self-evolving robotic agents in open, dynamic environments.

\bigskip

\textbf{Date:} April 11, 2026

\textbf{Correspondence:} xumu.xm@alibaba-inc.com

\textbf{Project Page:} \url{https://github.com/amap-cvlab/ABot-Claw}

}

\begin{document}
\maketitle
\vspace{-4pt}

\begin{figure}[!h]
    \centering
    \vspace{20pt}
\includegraphics[width=0.6\linewidth]{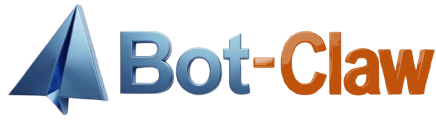}
    \label{fig:logo}
\end{figure}

\newpage
\tableofcontents
\newpage

\section{Introduction}
\label{sec:intro}

\begin{figure}[!b]  
    \centering  
    \includegraphics[width=1.0\linewidth]{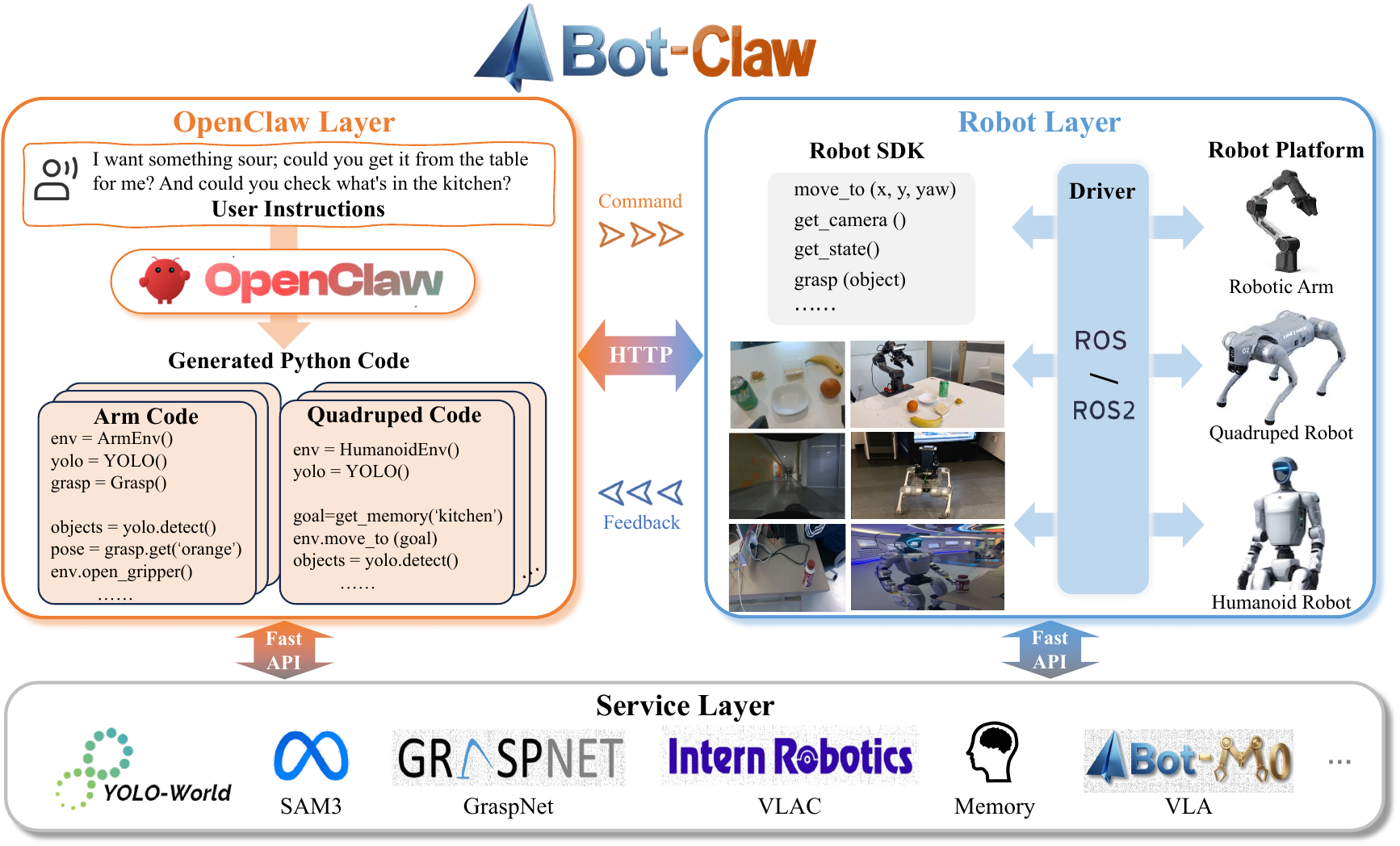}  
    \caption{\textbf{System Architecture of ABot-Claw.} Featuring a Layered Decoupling Strategy Among Interaction, Shared Services, and Embodiment Execution.  }  
    \label{fig:eng_figs}  
    \vspace{-1em}  
\end{figure}  

Recent robotic intelligent systems increasingly adopt a dual-system architecture inspired by cognitive science, often referred to as the System 1--System 2 paradigm \cite{kahneman2011thinking}. In this context, System 1 corresponds to fast, reactive modules that rely on learned priors to produce immediate responses. In contrast, System 2 represents a slower, deliberative process that supports structured reasoning, long-term planning, memory maintenance, and decision consistency across extended horizons. Current robotic intelligent systems commonly adopt architectures such as Vision-Language-Action (VLA) \cite{black2024pi_0,intelligence2025pi_,yang2026abot}, Vision-Language Navigation (VLN) \cite{chu2026abot,chen2025socialnav,zeng2025janusvln}, or World Action Model (WAM) \cite{li2026causal,ye2026world}, to enable fast perception and intuitive, reactive responses. These approaches demonstrate superior generalization in local tasks compared to traditional rule-based methods \cite{huo2025robust, zhu2014single, frazzoli2005maneuver}. However, such systems are fundamentally open-loop reactive agents, lacking explicit modeling of task states, environmental topology, and long-term memory. As a result, they struggle to support complex, long-horizon tasks that require continuous monitoring, dynamic adaptation, and coherent context maintenance. Simply enhancing the perceptual capabilities of System 1 cannot resolve the core limitations of missing memory, fragmented planning, and inconsistent execution \cite{huang2022inner}.

To overcome this bottleneck, researchers have proposed agent frameworks with advanced cognitive functions to implement System 2 capabilities \cite{liang2023code,fu2026cap,yokoyama2024vlfm}. These agents perform goal-driven reasoning through task decomposition, memory management, and self-reflection, while coordinating System 1 modules to execute concrete actions. This hierarchical integration of reasoning and reaction significantly improves behavioral organization and represents a critical step toward autonomous execution of complex tasks.

Nevertheless, most existing agents operate within closed sandboxes, relying on pre-registered toolkits and restricted to static pools of predefined functions \cite{schick2023toolformer}. When faced with novel actions such as greeting a guest through gesture or adapting to unexpected environmental changes, these systems often fail to generate solutions autonomously and must either abort or request human intervention \cite{huang2022language}. At a more fundamental level, such agents lack direct control over the operating system. They are unable to read or write files, invoke local applications, or maintain persistent processes, which results in a disconnection from the physical environment \cite{xie2023openagents}. Without the ability to interact meaningfully with real-world systems, they remain disembodied "brains in the air," unable to achieve true grounding in open-world environments \cite{driess2023palm}.

OpenClaw offers a key solution: as a self-hosted, locally running agent runtime, it possesses full system privileges, enabling shell command execution, GUI application control, event message listening, and unified integration across communication platforms such as WhatsApp, Telegram, and iMessage. With OpenClaw, AI agents can finally "take action" in the real world, achieving round-the-clock cross-platform autonomous operation. Crucially, OpenClaw supports natural language–driven dynamic skill evolution, allowing agents to generate code, validate logic, and deploy new functionalities autonomously, forming a closed-loop mechanism from intent understanding to capability realization. It is more than a communication channel; it is an infrastructure that endows AI with embodied presence.

While OpenClaw provides a powerful execution foundation that greatly enhances the flexibility and scalability of the perception-decision-action loop, it still lacks a corresponding high-level control architecture. As a result, it cannot natively support complex, long-duration, multi-device collaborative embodied tasks. The original framework does not clearly delineate the roles between System 1 and System 2, making sustained decomposition, monitoring, and correction of high-level intentions difficult. Its reliance on textual logs fragments visual, linguistic, and state information into multimodal silos, limiting contextual coherence \cite{gu2023rt}. Furthermore, the system lacks mechanisms for long-term state tracking and closed-loop feedback, making it prone to cascading failures under environmental perturbations. Additionally, integrating heterogeneous devices requires tight-coupled configurations, constraining dynamic collaboration and parallel execution.

In response, We propose \textbf{ABot-Claw}, built upon OpenClaw as the foundational runtime engine, to integrate three core technical innovations. First, it introduces a \textbf{unified embodiment interface and dynamic multi-agent scheduling mechanism} that connect heterogeneous robots through a shared skill layer, enabling capability-aware routing, parallel execution, and cross-embodiment collaboration. Second, it incorporates a \textbf{visual-centric multimodal memory} that stores object observations, place anchors, keyframes, and semantic visual representations in a shared memory space, allowing the runtime to retrieve persistent world context across time, space, and robot embodiments. Third, it equips execution with a \textbf{critic-based closed-loop feedback module}, instantiated with a generalist reward model \cite{zhai2025vision, tan2025robo}, to evaluate task progress online and support completion detection, local correction, or replanning under execution uncertainty. By tightly coupling high-level cognitive decision-making with low-level physical execution, ABot-Claw establishes an embodied intelligence framework capable of continuous learning, adaptation, and evolution in complex and dynamic environments, marking a significant step toward truly autonomous, general-purpose robotic systems.

\section{Method}
\label{sec3}

\subsection{Overview}
\begin{figure}[!t]  
    \centering  
    \includegraphics[width=1.0\linewidth]{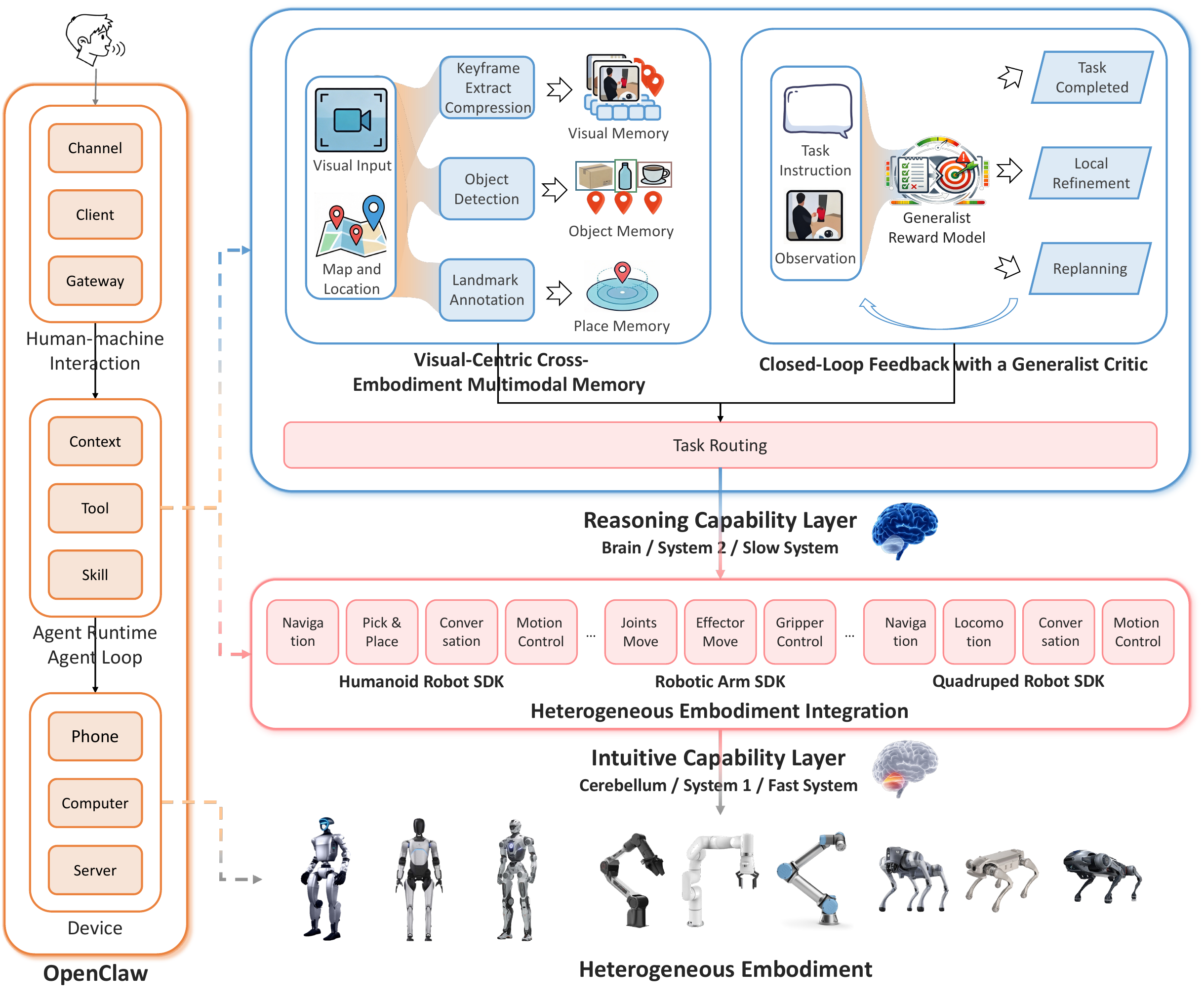}  
    \caption{\textbf{Overview of the ABot-Claw architecture.} Building upon the OpenClaw runtime, ABot-Claw enables persistent, cooperative, and self-evolving robotic agents through three core components: (1) a unified embodiment interface for coordinating diverse robot platforms, (2) a visual-centric multimodal memory for long-term, cross-embodiment context retention, and (3) a critic-based closed-loop feedback mechanism that supports online progress evaluation, local refinement, and dynamic replanning. Together, these components close the loop between high-level intent and low-level action in open, dynamic environments.}  
    \label{fig:overview-method}  
    \vspace{-1em}  
\end{figure}

Our goal in ABot-Claw is not to build an standalone embodied agent framework from scratch. Instead, we extend the OpenClaw runtime, originally designed for high-level software operation and task orchestration, into a general-purpose embodied runtime for real-world environments. This shift raises three practical challenges. First, the runtime must organize and control heterogeneous robot embodiments with very different actuation and control interfaces. Second, it must maintain persistent spatiotemporal context so that the agent can localize, recall, and act upon observations collected over long horizons. Third, it must remain robust under execution uncertainty, where open-loop plans often fail due to perception noise, environment changes, or control drift.

To address these challenges, we build ABot-Claw around three closely related components, as shown in Figure ~\ref{fig:overview-method}. We first extend OpenClaw with a unified embodiment interface that connects multiple robot types and exposes their capabilities through a shared skill layer. We then introduce a visual-centric multimodal memory that stores place anchors, object observations, semantic images in a shared memory space, so that the agent can reason over grounded environmental context rather than only the current observation. Finally, we incorporate a critic-style feedback module based on a generalist reward model, which provides an explicit signal of task progress during execution and supports termination, local correction, or replanning when needed. Together, these components turn ABot-Claw into an embodied runtime that can coordinate multiple robots, retrieve persistent world context, and react more robustly to real-world uncertainty. In the rest of this section, we describe these three components in turn and then summarize how they interact during end-to-end execution.

\subsection{Heterogeneous Embodiment Integration}
\label{sec:embodiment}

We begin by extending OpenClaw beyond a single physical platform. Many real-world tasks naturally span multiple workspaces and require complementary capabilities such as mobility, manipulation, and embodied perception. This motivates our one-runtime, multiple-bodies design, in which a single decision-making runtime coordinates a distributed pool of heterogeneous robot embodiments.

Under this design, ABot-Claw is not tied to one robot in one location. It instead acts as a centralized embodied agent that can perceive and intervene across multiple regions at the same time. In practice, this expands the set of tasks the system can solve and increases its capacity for parallel execution.

\textbf{Unified embodiment interface.} 
A central challenge in heterogeneous embodiment integration is that different robots expose fundamentally different low-level interfaces. A manipulator may rely on Cartesian planning from a fixed base, a mobile platform may expose navigation primitives, and a humanoid may require whole-body motion control. If these differences are passed directly to the reasoning layer, high-level planning becomes tightly coupled to hardware details and quickly becomes difficult to maintain.

To avoid this, we introduce a unified embodiment interface between the OpenClaw runtime and the underlying robots. Concretely, we provide ROS-based adapters for different robot types, and map their native functions into a shared set of callable skills. These skills capture intent-level actions such as navigation, observation, inspection, and manipulation primitives, while leaving the embodiment-specific execution to the corresponding local controller or model service. This design allows the high-level agent to issue actions through a common interface without reasoning directly about robot-specific command formats.

\textbf{Centralized runtime with multi-body execution.} 
At runtime, ABot-Claw maintains a dynamic pool of connected physical devices and acts as a coordination center. It tracks the availability, task progress, and basic status of each robot, and uses this shared state to organize execution across embodiments.

A practical benefit of this design is that a user request can be decomposed and dispatched across multiple robots when the subtasks are compatible with parallel execution. For example, when asked to prepare a reception environment, the runtime may send a mobile robot to inspect the corridor, assign another embodiment to approach the entrance area, and command a tabletop arm to arrange objects on a desk. Each robot executes the assigned subtask through its own local controller or service stack, while reporting execution status back to the shared runtime.

We note that ABot-Claw does not replace embodiment-side control. Low-level motion control, safety handling, and hardware-specific exception management remain on the robot side. Our contribution here is a runtime-level abstraction that allows OpenClaw to invoke and coordinate heterogeneous embodiments through a common task interface.

\textbf{Task routing and cross-embodiment collaboration.}
The value of multi-body embodiment is not limited to parallel execution of independent subtasks. In many practical scenarios, effective task completion requires handoff and cooperation across embodiments. In ABot-Claw, all robots operate as cooperative nodes that share both the same global task objective and the same environment memory, which allows the runtime to perform intent-aware routing and dynamic reassignment when a subtask exceeds the capability of a single robot or when cooperation yields better efficiency.

Our task allocation strategy jointly considers four factors: capability, location, load, and priority. Capability-based assignment matches task requirements to robot morphology, location-based assignment reduces travel time by favoring nearby robots, load-based assignment avoids overloading individual embodiments, and priority-based assignment ensures urgent requests are dispatched to available robots first. For instance, in an object transfer task, a mobile robot may deliver an object to a workstation, after which the runtime hands over the object pose and execution context to a fixed manipulator for precise placement or assembly. This centralized-brain, multi-body relay pattern allows ABot-Claw to solve compound spatial tasks that are difficult or inefficient for any single robot morphology alone.

While embodiment integration allows ABot-Claw to act through multiple bodies, these bodies still require a shared understanding of the environment. We therefore next introduce a memory system that aggregates observations across robots into a persistent, queryable, and action-ready world context.

\subsection{Visual-Centric Cross-Embodiment Multimodal Memory}
\label{sec:memory}

\begin{figure}[!t]  
    \centering  
    \includegraphics[width=1.0\linewidth]{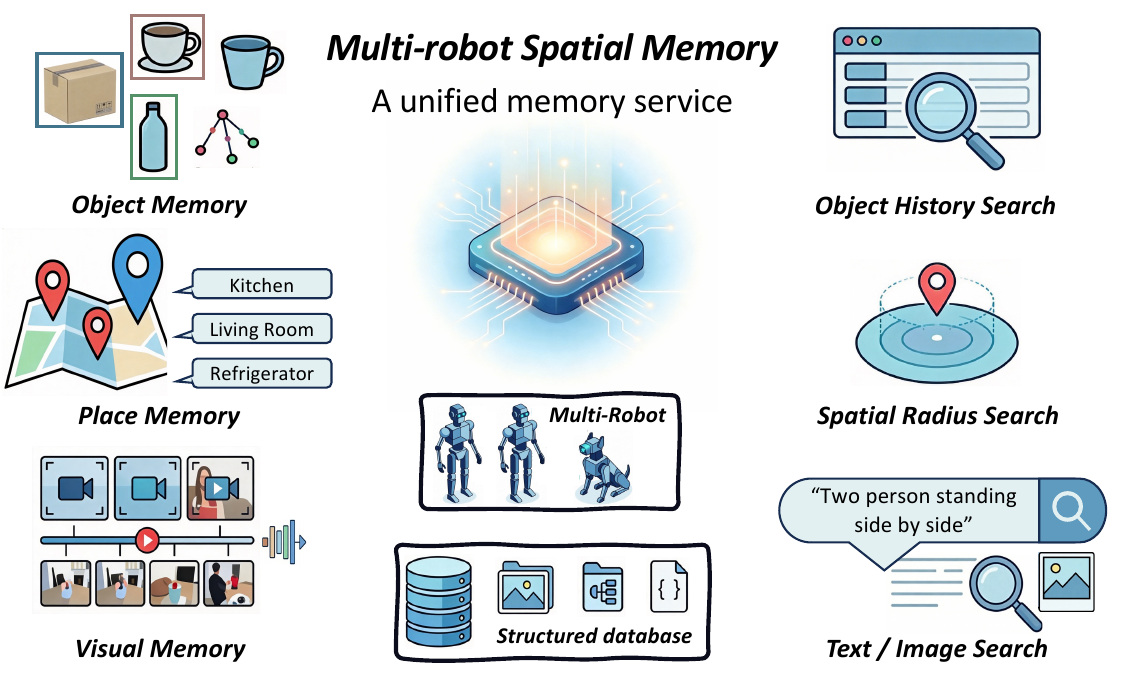}  
    \caption{\textbf{Visual-centric multi-robot memory.} ABot-Claw maintains a unified memory service that stores object memory, place memory, visual memory, and supports object history search, spatial radius search, text-based or image-based retrieval, and image-based retrieval. }  
    \label{fig:memory}  
    \vspace{-1.1em}  
\end{figure}  

Embodied execution requires more than instantaneous perception. A robot acting in the physical world must be able to remember where objects were previously seen, which places have already been explored, and what spatial relations are relevant to the current instruction. However, neither geometric maps nor text-only logs are sufficient on their own. Geometric maps offer precise localization but are poorly aligned with natural language, while textual summaries lose critical visual and spatial detail. This mismatch becomes more pronounced in open-world settings, where object categories, attributes, and relations are often long-tailed and difficult to compress into fixed symbolic labels.

To address this gap, we build a visual-centric multimodal memory that stores observations from different robots in a shared memory space. Rather than forcing all perception outputs into text, we preserve visual, semantic, and spatial information at a level that remains useful for later retrieval and execution. This memory serves as the grounding knowledge between high-level reasoning and physical execution: it gives the agent a persistent notion of where things are, what has been observed, and how the retrieved result can be translated into a concrete target for navigation or manipulation.

\textbf{Memory entities.}
We organize the memory around four types of entities, each corresponding to a different granularity of embodied context.

Visual memory stores scene-level observations together with their semantic and spatiotemporal context. Each entry contains a visual embedding extracted by a vision-language encoder, enabling open-vocabulary retrieval over unconstrained observations. In parallel, the system maintains sparse keyframes selected from long visual streams according to information density and scene novelty~\cite{apostolidis2022summarizing}. Each keyframe is associated with timestamp and pose metadata, allowing the runtime to retrieve not only semantically relevant scenes but also informative visual snapshots for environment initialization, historical review, and viewpoint revisiting across robots. By combining semantic embeddings with keyframe-level structure, visual memory supports both language-driven search and efficient trajectory summarization.

Object-centric memory anchors entities relevant to downstream interaction. Continuous visual detectors identify salient objects in the scene and store their category labels, observation timestamps, source robot identifiers, and associated spatial information like 3D poses. This memory is particularly useful for tasks such as grasping and placing, where the runtime needs object-level grounding and the ability to recover the last known observation of a target.

Place-anchor memory represents semantically meaningful locations in the environment. Through automatic registration or user annotation, selected coordinates are associated with names such as kitchen, entrance, or sofa area. These anchors discretize continuous space into language-friendly nodes, making it easier for the planner to reason about destinations, neighborhoods, and task-relevant regions.

\textbf{Retrieval mechanisms.}
Because different memory entities capture different kinds of context, we support two complementary retrieval paradigms.

For image-semantic memory, we adopt latent-space cross-modal retrieval. A vision-language encoder maps both visual observations and textual queries into a shared embedding space, and retrieval is performed by nearest-neighbor search using cosine similarity. Which allows the agent to search memory directly from natural language, even when the query refers to compositions of attributes and relations that would be difficult to specify through predefined tags alone.

For object-centric and place-anchor memory, we use structured retrieval over discrete metadata such as object category, source robot, time window, and spatial constraints. In practice, these structured filters can be combined with semantic retrieval. For example, the runtime may first retrieve semantically relevant frames and then narrow the result to observations from a recent time window or from a particular area of the environment.

\textbf{Navigable return protocol.}
A memory system is only useful for embodied execution if its outputs can be consumed directly by downstream controllers. To avoid burdening the language model with modality-specific parsing logic, we normalize all retrieval results into a unified navigable return protocol. The guiding principle is simple: multimodal input, spatialized output.

Regardless of whether a retrieval result originates from vector similarity over semantic frames or structured lookup over object detections, the memory module returns a standard action-ready representation, which includes semantic category, confidence score, visual evidence, and most importantly a stable 3D pose in the global coordinate frame. The runtime therefore can directly pass the returned pose into a navigation stack or motion planner for the selected robot embodiment.

\textbf{Shared memory across embodiments.}
All connected robots contribute to and consume from the same memory space. This shared design reduces repeated exploration, since newly added robots can reuse observations already gathered by other embodiments. In practice, this improves efficiency in multi-robot settings and allows the runtime to accumulate environmental context over time, even when the active embodiment changes from one stage of the task to the next.

Once such a shared memory is in place, the runtime can reason over persistent environmental context rather than reacting only to immediate observations. However, memory and planning alone are still insufficient for robust real-world execution. The next component of ABot-Claw therefore introduces an explicit critic that evaluates ongoing progress and closes the loop between planning and action.

\subsection{Closed-Loop Feedback with a Generalist Critic}
\label{sec:feedback}

Even with heterogeneous embodiments and grounded memory, real-world execution remains uncertain. Objects may move, detections may fail, manipulators may drift, and a plan that looked valid at the start of a trajectory may stop being appropriate after only a few steps. To improve robustness, we equip ABot-Claw with an explicit critic module that evaluates task progress during execution and provides decision signals for intervention.

\textbf{State evaluation with a generalist reward model.}
We instantiate the critic with a generalist reward model that takes the task instruction together with the current observation and returns a scalar progress signal. Intuitively, this score reflects how well the current state aligns with the intended goal. Unlike a binary success detector, such a signal can be useful throughout execution, providing an additional source of evidence about whether the current behavior is making progress.

\textbf{Adaptive correction and strategy switching.}
We use this critic signal to support three basic decisions during runtime. When the score exceeds a task-dependent threshold, the system marks the current subtask as complete and proceeds to the next stage. When the score remains below the completion threshold but still indicates improvement, the system keeps the current strategy and attempts local refinement, such as adjusting a target pose, updating the viewpoint, or repeating a short action sequence. When the score stagnates or drops significantly, the runtime treats the current strategy as ineffective and triggers replanning, potentially with help from the shared memory.

\textbf{Closing the execution loop.}
The critic does not only improve immediate task robustness, it also enriches the system's long-term experience. We log execution traces together with critic scores and feed them back into the runtime as structured experience. Over time, this creates a closed loop from execution to evaluation and from evaluation to future decision-making. In this sense, ABot-Claw extends OpenClaw from a task orchestration framework into a real-world runtime with online supervision and self-correction.

\subsection{End-to-End Execution Flow}

We now summarize how embodiment integration, memory, and feedback interact during runtime. Although the exact execution path depends on the task and selected robot, the overall behavior follows a common loop.

\textbf{Instruction grounding and embodiment selection.}
Given a user instruction, the OpenClaw runtime first interprets the task at a high level and selects a suitable embodiment or set of embodiments based on available skills, current status, and spatial context. If the task refers to previously observed places or objects, the runtime can query memory before acting, so that execution starts from grounded context rather than blind exploration.

\textbf{Memory-assisted action generation.}
Once the active embodiment is selected, the runtime invokes the corresponding skill or model service to generate executable actions. Depending on the task, this may include navigation, grasping, object search, or other embodied operations. At this stage, the memory layer may provide place anchors, object observations or semantic keyframe context that helps localize the target and reduce unnecessary search.

\textbf{Execution and progress evaluation.}
The selected robot then executes the action through its own control stack while the runtime monitors intermediate observations and status updates. During or after execution, the critic module evaluates task progress with respect to the original instruction and current observation, returning a signal that reflects whether the subtask is advancing toward completion.

\textbf{Update, refinement, or replanning.}
Finally, the runtime uses this feedback to decide what to do next. Successful executions are written back into memory when appropriate. Partially successful attempts may trigger local refinement. Failed or stalled attempts may trigger a new memory query, a revised plan, or a different embodiment assignment. This closes the loop between environment grounding, physical execution, and high-level task organization.

Overall, ABot-Claw integrates heterogeneous robot access, visual-centric multimodal memory, and critic-based execution feedback into a unified embodied runtime. This design gives OpenClaw a practical path from software task orchestration to real-world embodied operation, while keeping the system modular enough to support different robot morphologies, perception services, and control backends.

\section{Engineering Implementation}  
\label{sec:eng_imp}  
\subsection{Overview}
Robotic intelligence systems often integrate multiple heterogeneous components, including natural language interaction, task planning, low-level control, environment perception, reasoning, memory retrieval, and model service deployment. When these components are tightly coupled within a single runtime, unclear module boundaries are introduced. High upgrade costs are incurred. System portability is limited. The risk of single points of failure is also increased.

To overcome these challenges, a modular framework is designed. The system is decomposed into three decoupled modules with bidirectional communication, as shown in Figure~\ref{fig:eng_figs}. The first module is the OpenClaw interaction and scheduling layer. The second module is the robot embodiment execution layer. The third module is the shared service layer. In the proposed framework, high-level task understanding, capability querying, and task scheduling are handled by the OpenClaw layer. Execution logic that depends on specific hardware platforms and ROS interfaces is managed by the robot embodiment layer. Independent functionalities, such as perception, memory, and evaluation, are provided by the shared service layer. These functionalities are often computationally intensive or highly reusable.

By this layered decoupling strategy, each module is allowed to evolve independently through stable interfaces. System maintainability is improved. Scalability is also enhanced.

\subsection{OpenClaw Layer}

\begin{figure}[!t]  
    \centering  
    \includegraphics[width=1.0\linewidth]{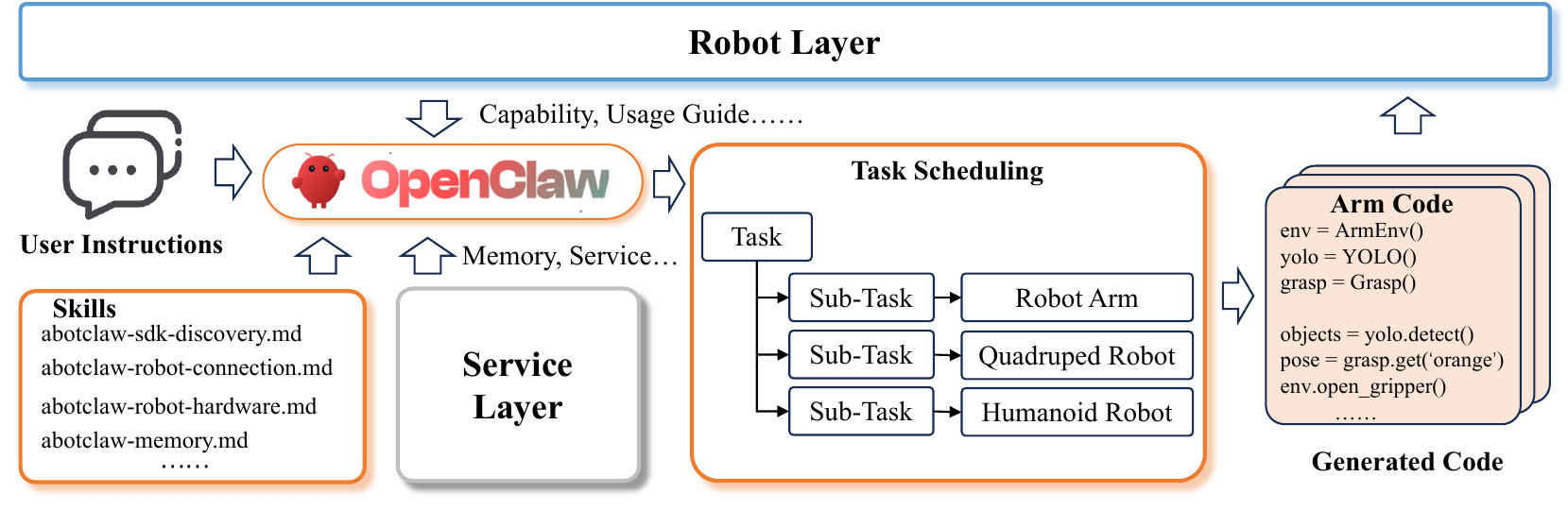}  
    \caption{\textbf{Workflow of OpenClaw Layer}. After receiving a user instruction, OpenClaw first loads the available skills. The robot layer then acquires robot-specific information and invokes service-layer modules, such as memory and process supervision. Based on the current observations, the user request is decomposed into a sequence of subtasks, from which executable Python code is generated and dispatched to the robot. During execution, OpenClaw continuously monitors robot states and task progress, enabling timely correction of execution errors.}  
    \label{fig:workflow_openclaw}  
    \vspace{-1em}  
\end{figure}  

The OpenClaw layer is designed as a unified interface for human--robot interaction and high-level decision making. Natural language instructions are parsed into structured task representations. Task context, skill abstractions, and device states are maintained within this layer. OpenClaw is not bound to any specific robotic platform. It functions as a general intelligent agent that abstracts and orchestrates heterogeneous robot and service resources as shown in Figure \ref{fig:workflow_openclaw}.

Task execution is formulated as a mapping from structured task specifications to available resources. Tasks are decomposed into sequential sub-tasks when necessary. Resource allocation is determined based on task requirements. For instance, fixed-base manipulation tasks are assigned to robotic arms, while large-area perception tasks are assigned to mobile platforms. This mapping ensures consistent and fine-grained scheduling without relying on platform-specific logic.

A capability-driven scheduling mechanism is implemented. Each robot exposes a structured description of its capabilities. Task allocation is performed by matching required capabilities with available resources. Static binding to robot identities is avoided. OpenClaw supports two scheduling modes: tasks can be assigned explicitly to user-specified robots, or resources can be selected automatically. This design improves flexibility and robustness under resource variation and system heterogeneity. New robots can be integrated through capability registration without modifying the scheduling policy.

Critically, OpenClaw outputs a fully executable Python script corresponding to the scheduled task. This script encodes all selected skills, resource bindings, and execution sequences. The generated Python file serves as the concrete interface between high-level task planning and robot-side execution, ensuring reproducibility and traceability of the planned operations. Simultaneously, the execution results are provided in real-time, and new executable code is output.

System-level coordination is also maintained. Task commands are dispatched, and execution feedback is collected. Failure handling and task replanning are triggered when required. External services, such as perception and memory modules, are accessed through standardized interfaces. Execution strategies are updated based on returned results. As a result, OpenClaw functions as a high-level orchestration module, producing executable task representations that bridge planning and physical execution.

A low-coupling integration scheme is employed. New robots and service modules are incorporated through standardized interfaces and capability descriptions. The core scheduling mechanism and OpenClaw output remain unchanged, improving scalability and supporting long-term system evolution.

\subsection{Shared Service Layer}
The shared service layer is designed to host computationally intensive and highly reusable modules that are weakly coupled with specific robot embodiments. Key functionalities, including object detection \cite{cheng2024yolo}, grasp perception \cite{fang2023anygrasp}, spatial memory, and task evaluation \cite{zhai2025vision}, are decoupled from both the robot layer and the OpenClaw layer. These functionalities are deployed as independent services. A unified interface is thus provided for capability access and extension.

The shared service layer is organized into three categories. First, perception services are provided, including object detection, depth-assisted localization, and grasp candidate generation. These services support environment understanding and action priors. Second, memory services are implemented to maintain shared spatial-semantic representations. Objects, locations, and scene structures are stored in a persistent form. This design supports context recovery and long-horizon task execution. Third, evaluation services are included to assess task progress and completion. Structured feedback signals are generated to support policy adjustment and execution monitoring.

The service layer is further designed as a platform for model-level extension. Advanced models such as VLA , VLN, and WAM, can be integrated through standardized interfaces. These models are typically computationally expensive and require independent runtime environments. Their deployment in the service layer avoids intrusive modification of robot control pipelines. A unified access interface is maintained for both OpenClaw and robot embodiments. This design enables a smooth transition from modular functionality to model-driven capability evolution.

Service-oriented deployment provides several engineering advantages. Computational resources, such as GPUs, can be centrally managed. Model deployment and version control are simplified. New models or algorithms can be incorporated through interface extension and service registration. No modification of the control logic or scheduling framework is required. This property supports rapid iteration and comparative evaluation of different methods.

The shared service layer also enables unified interaction across system components. OpenClaw can invoke services for task decomposition, environment understanding, and execution evaluation. Robot embodiments can access services during execution for online perception and feedback. Service outputs can be written back to shared memory or task context. A closed-loop interaction is thus established among OpenClaw, robot embodiments, and service modules.

In multi-robot scenarios, the shared service layer provides a foundation for information sharing and capability reuse. Perception, memory, and evaluation services are shared across robots. Redundant deployment is avoided. Environment knowledge can be accumulated and reused. For example, observations collected by a mobile robot can be stored and later accessed by a manipulation robot. This mechanism supports the transition from parallel execution to collaborative operation.

\subsection{Robot Layer}
The robot embodiment execution layer maps high-level action semantics into executable commands for specific hardware platforms. A ROS-based interface is adopted to standardize communication \cite{furrer2016robot}. This layer directly interfaces with physical systems and concentrates hardware-dependent components, including drivers, sensors, and control interfaces.

Robot-side functionalities are organized into shared and platform-specific modules. Shared modules define unified representations for execution flow, state feedback, and cross-layer communication. Platform-specific modules implement driver adaptation, topic mapping, and sensor integration. Embodiment heterogeneity is confined to these modules, while upper-layer interfaces remain unchanged. This design isolates hardware variation without introducing excessive abstraction.

ROS communication is used to achieve interface consistency across different robot. Although implementations differ, their interaction patterns are reduced to a set of stable topic structures. For each robot,  sensory inputs and robot states are expressed through standardized topics. System migration across platforms is achieved by adjusting configuration and topic mapping only. Upper-layer logic is not modified. Cross-embodied generalization is thus achieved under weak interface constraints.

Low-level control is not standardized across robot types. Instead, structural and functional differences are explicitly maintained within this layer. Each robot embodies specific physical capabilities, such as precise manipulation, human-centered interaction, or mobility over uneven terrain. Consequently, this layer serves as the concrete instantiation of embodied abilities. High-level scheduling in OpenClaw operates on these capability-defined execution units rather than on generic or abstract devices.

This layer also forms the execution interface of the system loop. High-level commands are received, and execution feedback is returned. Service-layer modules are invoked when additional perception or planning is required. Intermediate results are propagated to support task continuation. As a result, high-level task representations are consistently grounded into physical execution.
\section{Results Demonstration}
\label{sec:result}
\subsection{Overview}
To validate the effectiveness and generality of the proposed ABot-Claw system, experiments are conducted on three heterogeneous robotic platforms, including the Unitree G1 humanoid robot, the Unitree Go2 quadruped robot, and the Agilex Piper robotic arm. These embodiments differ substantially in morphology, mobility, and task affordances, providing a diverse testbed for assessing whether a unified runtime can generalize across robot types and task settings.

All tasks are specified in natural language, without predefined command templates or structured APIs. Given a user request, ABot-Claw interprets the instruction, selects appropriate embodiments and skills, and coordinates perception, planning, and control to execute the task. The following demonstrations examine this end-to-end capability across manipulation, mobile manipulation, cross-embodiment coordination, and guided navigation.

\subsection{Robotic Arm Demonstration}

\subsubsection{Interactive Search under Partial Observability}

\begin{figure}[!t]  
    \centering  
    \includegraphics[width=1.0\linewidth]{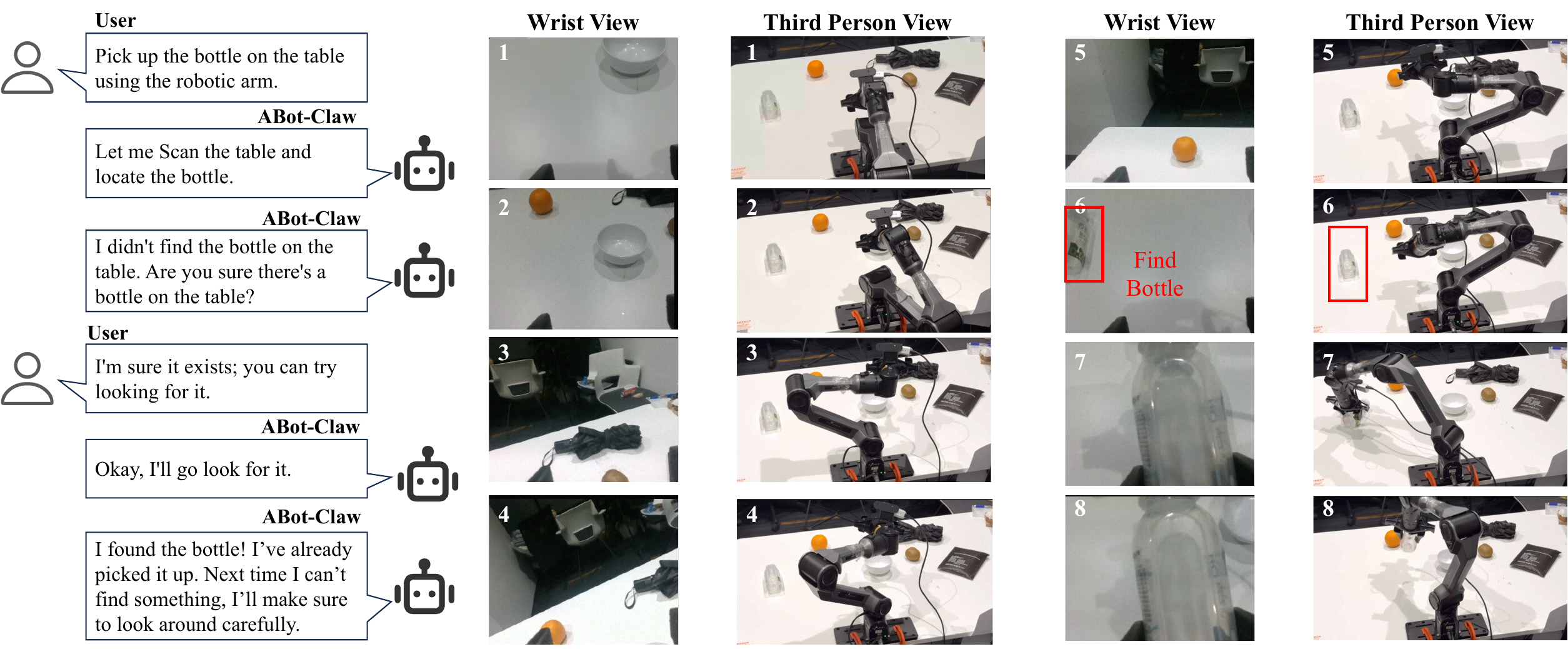}  
    \caption{\textbf{Demonstration of grasping an object outside the manipulator's initial field of view.}
    The bottle is not observed at the beginning. After a clarification query to the user, the manipulator searches the tabletop and successfully grasps the target object. Processes 1 to 5 represent the robotic arm's search process; the target object was found in process 6.}  
    \label{fig:armdemo1}  
    \vspace{-1em}  
\end{figure}  

We first consider a partially observable tabletop setting, where the requested bottle is not visible in the manipulator's initial view as shown in Figure~\ref{fig:armdemo1}. Rather than terminating early, the system identifies the uncertainty in the current observation and asks the user whether the object is absent or simply outside the field of view. Once the user confirms that the bottle is present, ABot-Claw generates a search procedure that combines viewpoint adjustment, repeated perception, and conditional branching.

The manipulator then incrementally expands its observable workspace until the target is detected, after which the system refines the target observation and executes grasping. This example highlights that the search behavior is not hard-coded as a fixed routine. Instead, it is synthesized online from the task context and updated based on user feedback and visual observations, illustrating the system's ability to act under partial observability.

\subsubsection{Manipulation under Ambiguous Instructions}

\begin{figure}[!t]  
    \centering  
    \includegraphics[width=1.0\linewidth]{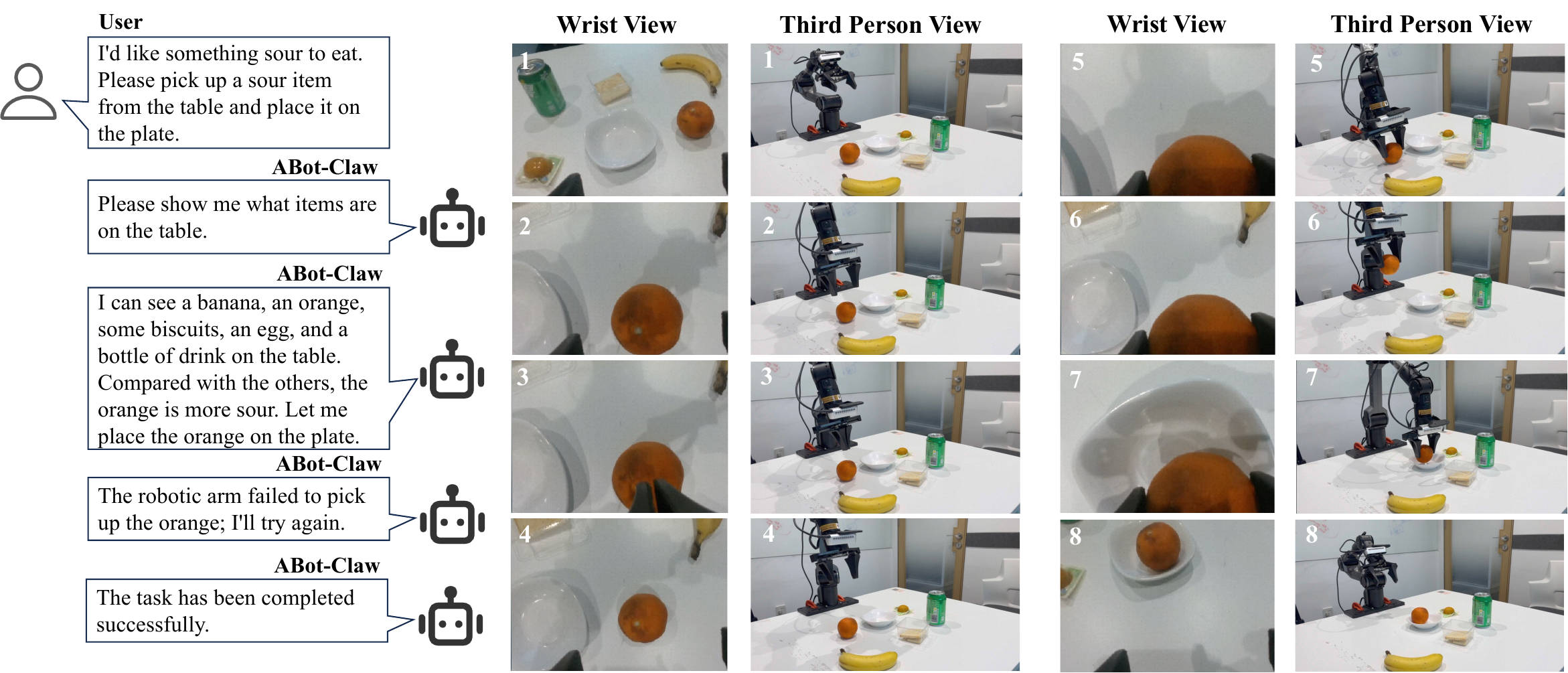}  
    \caption{\textbf{Demonstration of semantic manipulation under ambiguous instructions.} 
    Given an abstract user request, the manipulator interprets the semantic intent, selects the appropriate object, and completes the grasp-and-place task.}  
    \label{fig:armdemo2}  
    \vspace{-1em}  
\end{figure} 

We next study a manipulation task specified by a semantic attribute rather than an explicit object name. The instruction asks the robot to pick \textit{``something sour''} from the table and place it on a plate as shown in Figure~\ref{fig:armdemo2}. To solve this task, the system first perceives the scene and enumerates candidate objects in the workspace, then performs semantic reasoning to determine which item best matches the requested property.

After selecting the target, ABot-Claw generates and executes the corresponding manipulation program for grasping and placement. The result shows that the system can ground ambiguous, high-level language into concrete object selection and motor behavior without requiring fully specified object-level commands. In addition, when the initial grasp fails, the monitoring module detects the unsuccessful outcome and triggers replanning, recomputing the grasp pose for a new attempt. This demonstrates that execution is not purely open loop, but continuously monitored and corrected online.

\subsection{Humanoid Robot Demonstration}

\subsubsection{Mobile Manipulation for Object Delivery}

\begin{figure}[!t]  
    \centering  
    \includegraphics[width=1.0\linewidth]{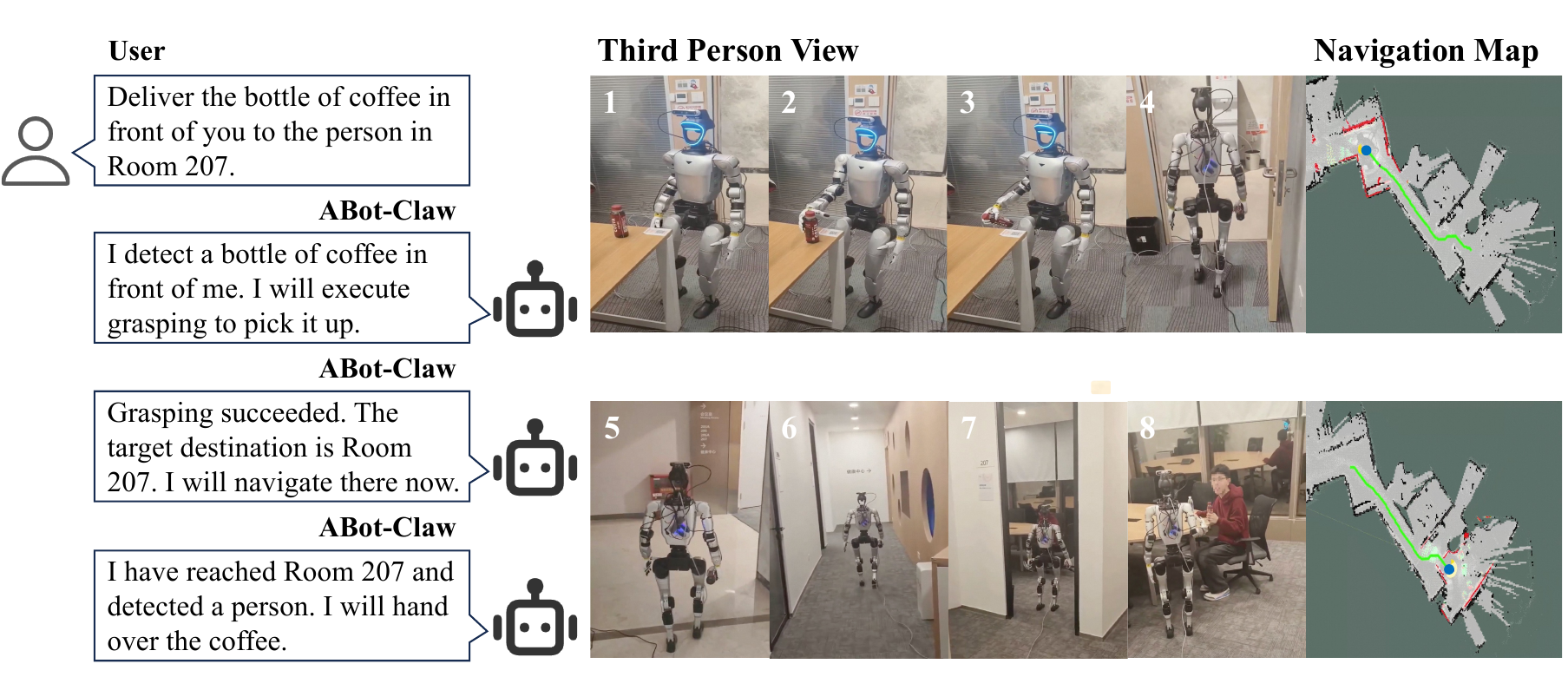}  
    \caption{\textbf{Demonstration of humanoid mobile manipulation for delivery.} 
    Given the instruction to deliver the coffee bottle in front of the robot to the person in Room 207, the humanoid grasps the object, navigates to the target room, and hands it to the recipient.}  
    \label{fig:humanoiddemo2}  
    \vspace{-1em}  
\end{figure}

Figure~\ref{fig:humanoiddemo2} presents a long-horizon humanoid mobile manipulation task: delivering the coffee bottle in front of the robot to a person in Room 207. This task couples local object interaction with environment-scale navigation and person-aware delivery.

ABot-Claw first grounds the object reference from the current visual observation and invokes a grasping skill to pick up the coffee bottle. After the grasp result is verified, the destination is grounded as a navigation goal associated with Room 207, and the humanoid is dispatched toward the target room. Upon arrival, the robot detects the person at the destination and completes the task through a handover action. This example demonstrates coordinated perception, manipulation, navigation, and interaction within a single execution loop.

\subsubsection{Cross-Embodiment Task Reassignment for Robot Inspection}

\begin{figure}[!t]  
    \centering  
    \includegraphics[width=1.0\linewidth]{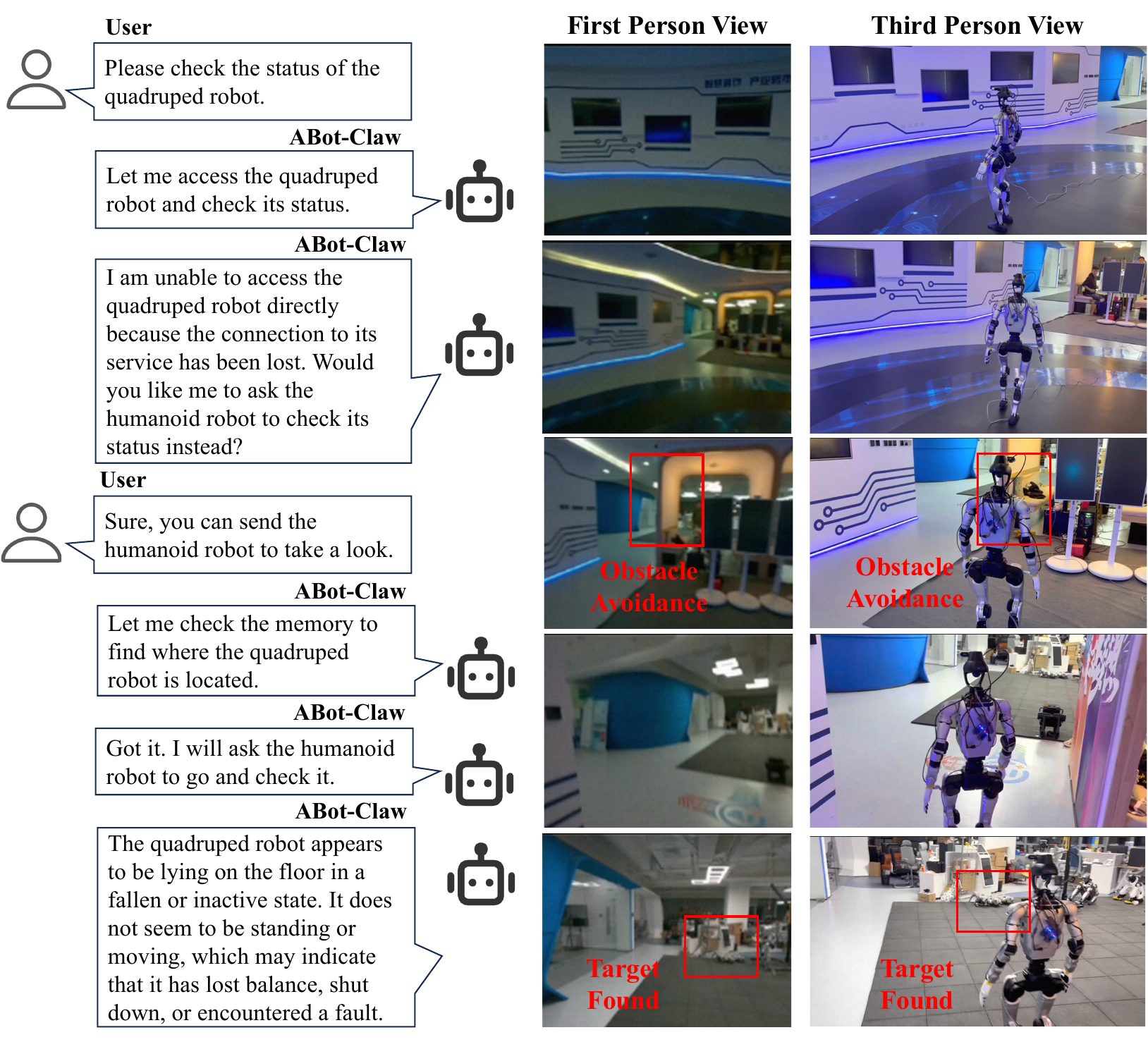}  
    \caption{\textbf{Demonstration of humanoid-assisted inspection.} 
    When the user requests the status of the quadruped robot, the system first detects that the quadruped is disconnected. It then queries the memory module to retrieve the robot's last known location and dispatches the humanoid robot to inspect its condition.}  
    \label{fig:humanoiddemo1}  
    \vspace{-1em}  
\end{figure}

We next consider a cross-embodiment inspection scenario in which the user asks for the status of a quadruped robot as shown in Figure~\ref{fig:humanoiddemo1}. The system first attempts to access the quadruped directly, but the service is unavailable and direct communication fails. Instead of terminating, ABot-Claw queries the shared memory to recover the quadruped's last known location and uses that information to reassign the task.

The humanoid is then selected as an alternative embodiment and dispatched to the target area for physical inspection. It navigates through the environment, observes the quadruped, and reports the resulting status. This case illustrates how shared memory and embodiment scheduling allow the system to recover from robot-side failure by transferring the task to another available platform.

\subsection{Quadruped Robot Demonstration}
\subsubsection{Quadruped-Guided Visitor Reception}

\begin{figure}[!t]  
    \centering  
    \includegraphics[width=1.0\linewidth]{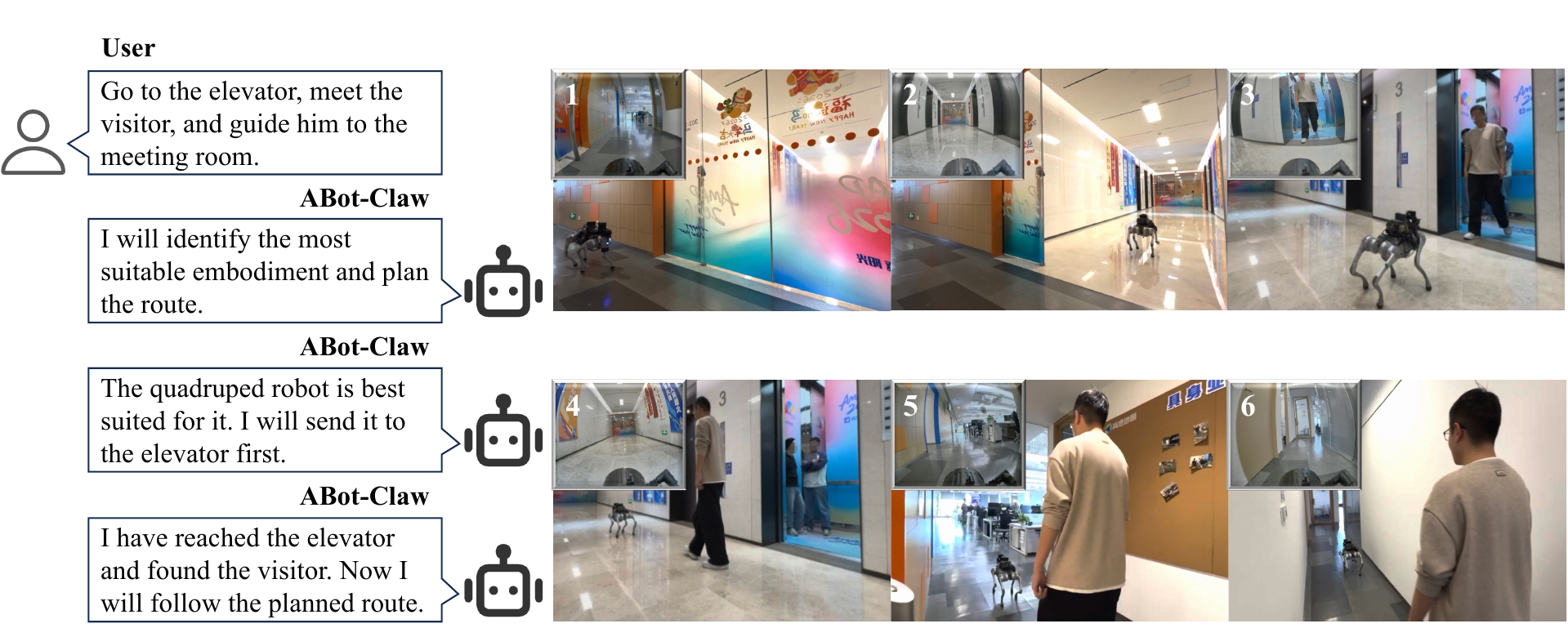}  
    \caption{\textbf{Demonstration of quadruped-guided visitor reception.} 
    Given the instruction to meet a visitor at the elevator and guide the visitor to the meeting room, the system selects the quadruped robot, dispatches it to the elevator, detects the visitor, and leads the visitor to the destination.}  
    \label{fig:quadrupeddemo1}  
    \vspace{-1em}  
\end{figure}  

The final example focuses on visitor reception and guidance using the quadruped robot. The task is to meet a visitor at the elevator and escort the visitor to the meeting room as shown in Figure~\ref{fig:quadrupeddemo1}. Compared with the previous tasks, this scenario emphasizes embodiment selection, agile navigation, person detection, and sustained guidance over a longer spatial range.

ABot-Claw selects the quadruped as the most suitable embodiment because the task primarily requires mobility and person-guiding behavior rather than manipulation. The robot is dispatched to the elevator area, where it searches for and identifies the arriving visitor. After detection, it initiates the guidance behavior and leads the visitor along the planned route to the meeting room while the runtime monitors task progress. This demonstration highlights that embodiment selection can be conditioned on task structure and that the same runtime can coordinate reception-style behaviors on a mobile legged platform.

\section{Conclusion}
\label{sec:conclusion}

In this report, ABot-Claw was presented as an embodied intelligence framework that extends the OpenClaw runtime into a general foundation for real-world robotic execution. To reduce the gap between high-level reasoning and physical interaction, three key components were integrated into the framework: 1) critic-based closed-loop feedback mechanism for online closed-loop feedback and policy adjustment, 2) a visual-centric multimodal memory for cross-modal context retention, and 3) an elastic multi-agent scheduling architecture based on dynamic routing tables. By adopting a decoupled engineering design, heterogeneous embodiment interfaces were standardized, which enabled coordinated deployment across diverse robotic platforms, including robotic arms, humanoids, and quadrupeds.

Experimental results showed that ABot-Claw can autonomously execute complex manipulation and search tasks. Robust performance was demonstrated under partial observability, and ambiguous semantic instructions were successfully grounded into executable actions. These results validate the effectiveness and flexibility of the proposed framework in open-world robotic settings.

Overall, ABot-Claw provides a scalable infrastructure for general-purpose embodied agents. It offers a practical step toward robotic systems that are more persistent, cooperative, and adaptive in real-world environments.

\clearpage
\section{Contributions and Acknowledgments}
\label{sec:contributions}
\setlength{\parskip}{0pt} 
\setlength{\itemsep}{0pt} 
\setlength{\parsep}{0pt}  
\raggedcolumns

\subsubsection*{Contributions}
Author contributions in the following areas are as follows:

\begin{itemize}
    \item \textbf{System Design \& Implementation:} Dongjie Huo, Haoyun Liu, Guoqing Liu, Dekang Qi, Yandan Yang, Xinyuan Chang, Feng Xiong
    \item \textbf{Humanoid Robot Deployment:} Zhiming Sun, Maoguo Gao(Deepblue College), Jianxin He(Deepblue College), Dekang Qi
    \item \textbf{Quadruped Robot Deployment:} Guoqing Liu, Dongjie Huo, Haoyun Liu
    \item \textbf{Robotic Arm Deployment:} Dongjie Huo, Haoyun Liu
    \item \textbf{Writing:} Dongjie Huo, Haoyun Liu, Dekang Qi
    \item \textbf{Project Lead:} Xinyuan Chang, Feng Xiong
    \item \textbf{Advisor:} Mu Xu$^\dagger$, Zhiheng Ma, Xing Wei
\end{itemize}





{\renewcommand{\thefootnote}{\fnsymbol{footnote}}\footnotetext[2]{Corresponding author: xumu.xm@alibaba-inc.com}}

\subsubsection*{Acknowledgments}
We thank DeepBlue College for their support in the humanoid robot deployment, providing access to the Unitree G1-Romp Edu robot and the LinkerBot-O6 hand system.

\clearpage

\bibliographystyle{plainnat}
\bibliography{main}

\clearpage
\beginappendix

\section{Appendix}

\let\clearpage\relax

\end{document}